\documentclass[conference]{IEEEtran}
\IEEEoverridecommandlockouts
\usepackage{cite}
\usepackage{amsmath,amssymb,amsfonts}
\usepackage{algorithmic}
\usepackage{graphicx}
\usepackage{textcomp}
\usepackage{caption}
\usepackage{subcaption}
\usepackage{xcolor}
\usepackage{todonotes}
\usepackage{framed}
\usepackage{float}
\usepackage[export]{adjustbox}

\usepackage[colorlinks=true, linkcolor=black,urlcolor=black, citecolor=black]{hyperref}
\def\BibTeX{{\rm B\kern-.05em{\sc i\kern-.025em b}\kern-.08em
    T\kern-.1667em\lower.7ex\hbox{E}\kern-.125emX}}
\begin{document}

\title{Analysing the Needs of Homeless People Using Feature Selection and Mining Association Rules \\
}

\newcommand{\newlineauthors}{%
  \end{@IEEEauthorhalign}\hfill\mbox{}\par
  \mbox{}\hfill\begin{@IEEEauthorhalign}
}

\author{\IEEEauthorblockN{José M. Alcalde-Llergo}
\IEEEauthorblockA{\textit{Artificial Vision Applications (AVA)} \\
\textit{University of Córdoba}\\
Córdoba, Spain \\
i72alllj@uco.es}
\and
\IEEEauthorblockN{Carlos García-Martínez}
\IEEEauthorblockA{\textit{Computing and Numerical Analysis} \\
\textit{University of Córdoba}\\
Córdoba, Spain \\
cgarcia@uco.es}
\and
\IEEEauthorblockN{Manuel Vaquero-Abellán}
\IEEEauthorblockA{\textit{Nursing, Physiotherapy and Pharmacology} \\
\textit{University of Córdoba}\\
Córdoba, Spain \\
en1vaabm@uco.es}
\and
\IEEEauthorblockN{Pilar Aparicio-Martínez}
\IEEEauthorblockA{\textit{Nursing, Physiotherapy and Pharmacology} \\
\textit{University of Córdoba}\\
Córdoba, Spain \\
n32apmap@uco.es}
\and
\IEEEauthorblockN{Enrique Yeguas-Bolívar}
\IEEEauthorblockA{\textit {Computing and Numerical Analysis} \\
\textit{University of Córdoba}\\
Córdoba, Spain \\
eyeguas@uco.es}
}

\maketitle

\begin{abstract}
Homelessness is a social and health problem with great repercussions in Europe. Many non-governmental organisations help homeless people by collecting and analysing large amounts of information about them. However, these tasks are not always easy to perform, and hinder other of the organisations duties. The SINTECH project was created to tackle this issue proposing two different tools: a mobile application to quickly and easily collect data; and a software based on artificial intelligence which obtains interesting information from the collected data. The first one has been distributed to some Spanish organisations which are using it to conduct surveys of homeless people. The second tool implements different feature selection and association rules mining methods. These artificial intelligence techniques have allowed us to identify the most relevant features and some interesting association rules from previously collected homeless data.
\end{abstract}

\begin{IEEEkeywords}
AI for inclusivity, Feature selection, Association rules, Homelessness, Data collection
\end{IEEEkeywords}

\section{Introduction}
According to the European Commission \cite{articleUE}, in 2019, more than 700,000 people were sleeping rough in the European Union, approximately 70\% more than in 2009. Homeless people are one of the most vulnerable groups in society, continuously exposed to accidents, diseases, violence, precarious nutrition and other factors that compromise their life. This problem tends to affect more to adult men \cite{womenHomeless}, but in recent years we are seeing a growth in homelessness among people between 15 and 29 in Europe \cite{articleUE}. 

Homeless people rely on non-governmental organisations (NGOs) to obtain essential supplies such as food, clothes, blankets or hygiene materials. In order to provide these supplies, NGOs need to collect information about the main homeless people requirements, so they have to hire people or look for volunteers to conduct surveys of this social group. However, some previous data collections contain information about homeless people. For example, the Spanish National Statistic Institute (INE) conducted different types of surveys on socio-demographic characteristics, housing, work activity, health, economic situation, studies, social services, and relationship with the judicial system of 3,433 homeless people \cite{encuesta}.

In recent years, the evolution of computer systems and artificial intelligence (AI) algorithms has made possible to develop instruments that could help to perform these data collection and analysis tasks. However, there are no methods in the literature that attempt to find interesting relationships between attributes of homeless people, nor do they attempt to predict the health status of a homeless person. The SINTECH project aims to develop a set of tools to address these tasks and facilitate the work of NGOs in helping homeless people.

We briefly summarise our contributions as follows:

\begin{itemize}
    \item We propose a mobile application to collect data on the homeless. This application provides other functionalities such as dissemination of NGOs information, locating homeless help centres, or contacting help services.
    \item We provide some data analysis AI tools to obtain interesting information about homeless people. These tools are divided in two modules: a feature selection module and an association rules generation module.
    \item We evaluate the results obtained by the different AI modules applied over the INE homeless people dataset \cite{encuesta}. The final results are composed of: a set of relevant features to predict the health condition of a homeless person, some classifiers to predict this health condition and a set of association rules which describe interesting relationships between some of the data attributes.
    
\end{itemize}

The rest of the paper is organised as follows. Section II surveys some
related work. In Section III we present a description of the mobile application and the applied AI methods. The experimental design and the results obtained by the different tools are analysed in Sections IV and V, respectively. Finally, we expose our final conclusions in Section VI.

\section{Related  Works}

\subsection{Study on Homelessness}

Due to the growing number of homeless people, many studies are being carried out to try to help this social group to improve their health and living conditions \cite{ant_st1, ant_st_COVID}. More specifically, this project was inspired by a study carried out on homeless people in the Spanish city of Cordoba \cite{alvaro}. During the study, the author analysed the main health problems suffered by homeless people in this city, as well as the relationship that they may have with different socio-demographic variables. Its final objective was to identify the basic needs of this social group and how to address them. Another study was conducted by the INE \cite{encuesta} to analyse the socio-demographic characteristics, housing, work activity, health, economic situation, studies, and discrimination towards homeless people in Spain.

\subsection{Mobile Applications}


Due to the widespread use of smartphones today, mobile applications have demonstrated to be one of the most powerful tools to promote the inclusivity of disadvantaged social groups. In the case of homelessness, an example is Shelter App \cite{shelterApp}, a mobile application whose objective is to help homeless and low-income families connecting to different resources and services. It provides the user with a list of help resources located on a map, so that by selecting them they can obtain the necessary information to contact them. However, this application is currently only operating in some of the USA states. 

\subsection{Artificial Intelligence}
In recent years, artificial intelligence has proven to be applicable with very good results to problems in all kinds of fields. We will focus our work in two popular kind of AI tools: feature selection \cite{FS} and mining association rules \cite{assoc_rules}.

Feature selection is a popular data preprocessing strategy that has demonstrated to be effective and efficient to prepare datasets for both data mining and machine learning problems. It is especially useful when dealing with high-dimensional problems, because it allows to reduce the complexity of the problem and the computational cost of applying different algorithms to it, while improving the accuracy of the generated models \cite{FS_Effectiveness}.

On the other hand, association rule mining is one of the most popular data mining techniques. It is traditionally used to find relationships between attributes of transactional datasets and generate rules that relate them. More recent studies propose new methods to optimise and apply association rules to all kind of problems, including classification. An example of this can be seen in \cite{reglasCarlos}, where the authors propose a new heuristic for generating class association rules in a medical dataset.

\section{Sintech Developed Tools}

In this section, an overview of the developed mobile application is briefly discussed, followed by the description of the AI modules which we have implemented. A simplified overview of all these tools is depicted in Figure \ref{fig:diagrama}.

\begin{figure}[!h]
\centerline{\includegraphics[width=0.49\textwidth]{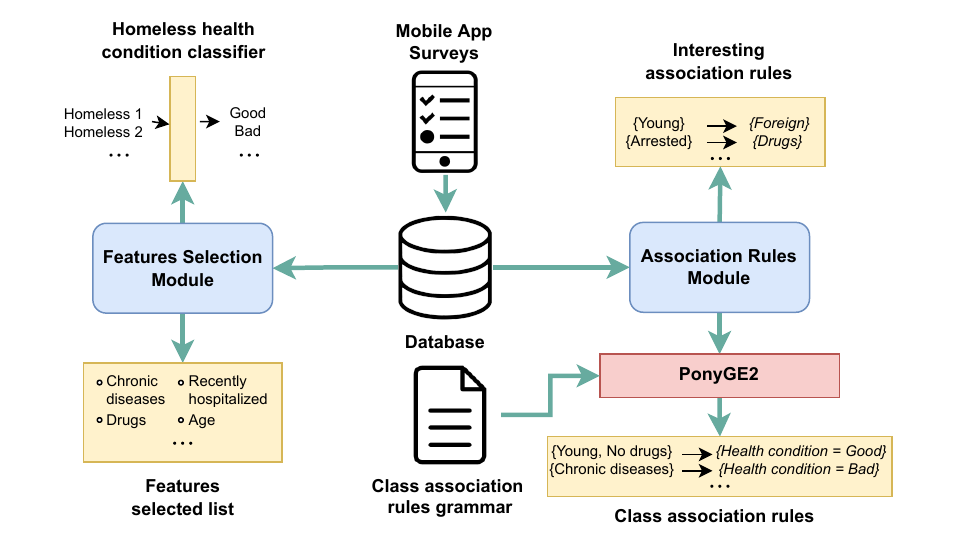}}
\caption{Block diagram of the main SINTECH tools.}
\label{fig:diagrama}
\end{figure}

\subsection{Mobile Application}
The main objective of the mobile application is to facilitate the task of collecting data on homeless people. The idea is that any citizen who has the app can carry out surveys of homeless people and send them directly to NGOs. The survey will contain multi-choice questions about daily life, health, work situation, studies, etc. This information will help NGOs to know the main needs of this social group, as well as being the future input for our AI models. However, data collection is not the only functionality of the application. The app also includes other functionalities such as a wall where the NGOs can post information about homeless people and how to collaborate with them; a quick contact system with the main help services which could assist a homeless person; and an aid points location map where we can find different places where NGOs provide services for homeless people. The application will have a simple menu-driven interface so that anyone can use it.

\subsection{Feature Selection}
The feature selection module applies different features selection methods from the scikit-learn python package \cite{scikit-learn} to obtain the most important attributes related to a relevant characteristic. The applied feature selection methods can be classified traditionally as: filter, wrapper and embedded methods \cite{featureselectionTypes}. In the case of the former, we have applied different statistics in order to study the relationship between the attributes of a homeless person and the objective feature to predict (health condition). Secondly, as wrapper methods we have applied some sequential and recursive feature selection algorithms \cite{RFE_SVM}. From the last group we have applied an embedded method based on Random Forest. Finally, we have used as our last feature selector a Decision Tree, whose attributes selected for the first splits have been considered as more relevant features to predict the objective class.

\subsection{Association Rules}
The association rules module extracts some relationships between the different features of the surveyed homeless people in the form of a rule: \{Antecedent $\rightarrow$ Consequent\}. The objective is to find interesting rules that allow NGOs to discover the main needs of homeless people. In addition, we implemented a sub-module to find class association rules to predict the health state of a homeless by means of an evolutionary algorithm. Finally, we have also developed a trivial rule filter sub-module which automatically removes obvious association rules.

\section{Experiments}

This section will describe the different experimental processes that have been conducted by using the tools developed during this project. More specifically, the experimentation phase will be mainly divided into two parts, one focusing on the feature selection and the other on the association rules.

\subsection{INE Homeless Dataset}
The mobile application has been already distributed to different Spanish NGOs, but we have not yet collected enough homeless information. Therefore, to demonstrate the performance of the proposed tools, we conducted the feature selection and the association rules experiments over the previously collected INE homeless dataset \cite{encuesta}. The questions of the INE survey are not the same as those collected by the application but in both cases they are multi-choice or numerical questions. The INE dataset is composed of the responses to 283 questions from 3,433 homeless people. After a preliminary preprocessing step, 46 questions were considered irrelevant because of the large number of people who did not answer them, resulting in 187 features for our final dataset. Examples of these variables are age, health status or education.

\subsection{Feature Selection Experimentation}
For this part of the experimentation, the 3,433 collected surveys were randomly divided in a stratified way (according to the health state variable) into training and test sets, accounting for 70\%, and 30\% of the original dataset, respectively. A Random Forest classifier was selected to measure the performance of the different feature selection algorithms, due to its easy applicability and the fact that it can be used as a feature selector on its own. Moreover, we have applied two different encodings to transform categorical variables into numerical ones to achieve the correct performance of the applied algorithms: an ordinal encoding has been applied to those characteristics whose values imply an order among them, and a 1-of-K encoding (also known as one-hot encoding) has been applied to the rest of the categorical variables. This encoding resulted in 632 final variables as input for the feature selection methods, which makes the task of feature selection even more relevant.

During these experiments, we compared the performance of 8 different feature selectors: three filter methods based on the statistics $chi^2$, mutual information and F-statistic; a forward sequential feature selector (FFS); two variants of recursive feature elimination (RFE and RFECV); an embedded feature selector based on Random Forest; and a Random Forest classifier as selector by itself. The documentation of these algorithms can be found in \cite{scikit-learn}. This comparison has been based on the Correct Classification Rate (CCR) metric (\ref{eq:ccr}) obtained by a Random Forest classifier over the test set by using only the selected features for each different selector. For a fair comparison, six has been considered as the final number of characteristics for filter methods, FFS and RFE. This number has been selected after studying the optimum number of features to predict the objective variable by using the different algorithms. The rest of the applied algorithms automatically detect their optimal number of features. In addition, the most relevant features to predict the objective variable, the homeless health state, will be obtained according to the number of times they appeared in the top six variables selected by the different methods.

\begin{equation}
    CCR = \frac{Correctly\:classified\:patterns}{Total\:number\:of\:patterns}
    \label{eq:ccr}
\end{equation} 

\subsection{Association Rules Experimentation}
To perform the association rules mining we need a dataset composed by transactions. For this purpose, we discretised all the numeric variables of our dataset, and then we applied a 1-of-K encoding to obtain a transactional format. The association rules mining will be performed over the whole transaction dataset by using two different strategies. First of them will use the Apriori \cite{apriori} and FP-growth \cite{fpgrowth} algorithms to obtain the frequent itemsets based on the support, or frequency of occurrence, of the possible itemsets. Then, the rules will be selected by the rules generator algorithm based on the confidence metric (\ref{eq:confidence}). All the methods used for this first approach are implemented in the mlxtend python library \cite{mlxtend}. After that, we will apply a pre-filtering based on other metrics to measure the interest of the rules, such as lift (\ref{eq:lift}), antecedent support and consequent support. The lift will allow our algorithm to take into account not only the antecedent support but also that consequent support, while these supports individually will show if the high confidence of a rule is due to the high occurrence of their antecedent or consequent. Finally, these obtained rules will be automatically filtered by a trivial rules filter module, where the most obvious rules, based on the strongly related attributes of the dataset, will be removed to ease the search for interesting rules. 

\begin{equation}
    Confidence = \frac{Support(X,Y)}{Support(X)}
    \label{eq:confidence}
\end{equation}
\begin{equation}
    Lift = \frac{Support(X,Y)}{Support(X)\cdot Support(Y)}
    \label{eq:lift}
\end{equation}
where X and Y are itemsets.

Our second strategy is based on PonyGE2 \cite{ponyge2}, which is a population-based evolutionary algorithm implemented in python. We have developed a new grammar that can be interpreted by this algorithm so that it can generate class association rules \cite{class_association_rules}. These rules will be generated based on a custom metric obtained by the product of the precision and the recall, using a population size of 500 and considering the health state of the homeless person as objective variable. This strategy has been considered very interesting due to the great potential of these algorithms for optimisation and the flexibility that the grammar brings to the rules.

\section{Analysis of Results}
In this section, we present and discuss some of the most interesting results obtained by the different tools developed in this work. We are going to divide this section into three different parts. In the former, we are going to talk about the developed mobile application. In the second one, we are going to analyse the results obtained by the features selection module. Finally, in the third subsection, we are going to study the generated association rules to obtain conclusions from them.

\subsection{Mobile Application}

Figure \ref{fig:app_fotos} shows some of the different functionalities of the developed mobile application (Intellectual Registration: RTA-1534-21). Its performance has been evaluated through diverse white-box tests during its development and various black-box tests after its deployment on different devices. Through these tests, it has been concluded that the application works properly and is ready for use by organisations and citizens.

\begin{figure}[h]
     \centering
     \begin{subfigure}[b]{0.14\textwidth}
         \centering
         \includegraphics[width=\textwidth]{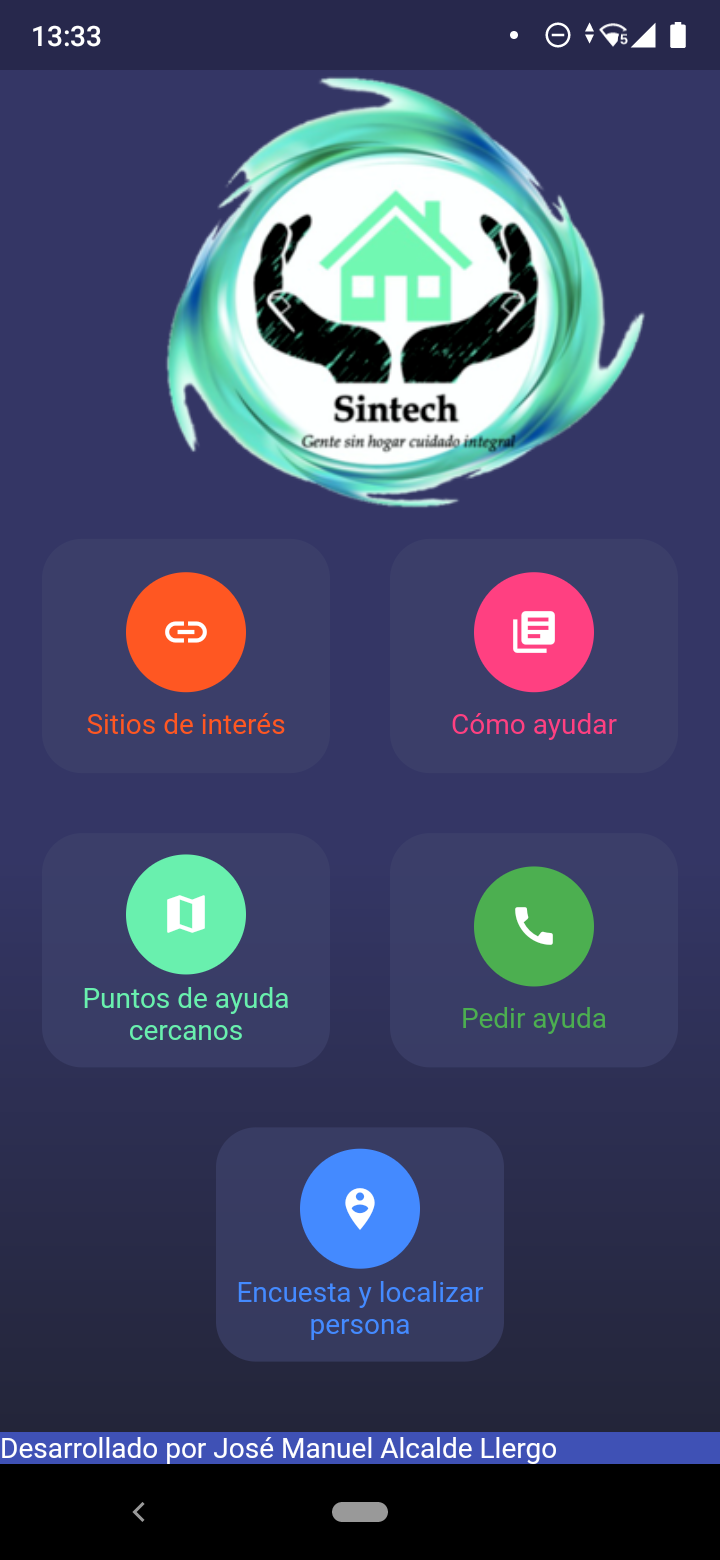}
     \end{subfigure}
     \hfill
     \begin{subfigure}[b]{0.14\textwidth}
         \centering
         \includegraphics[width=\textwidth]{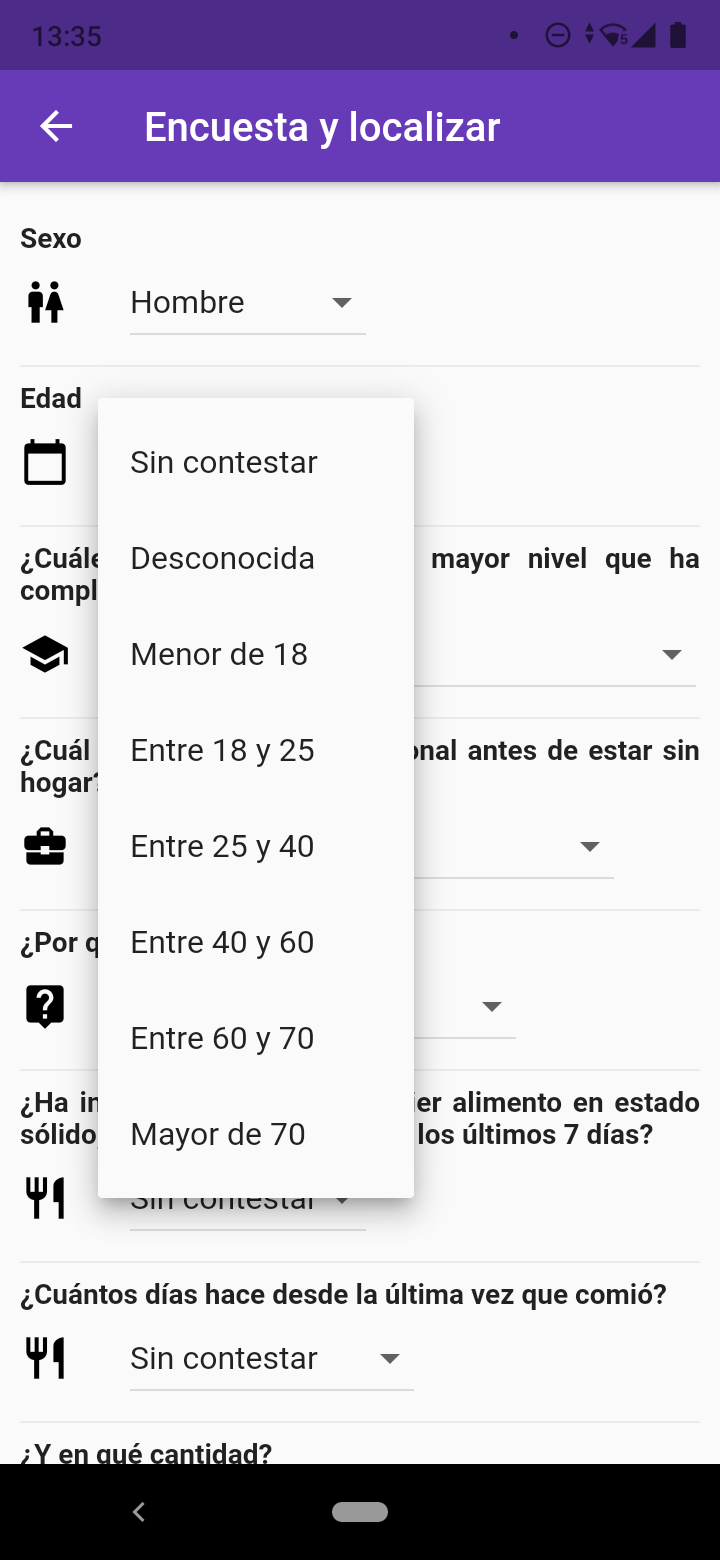}
     \end{subfigure}
     \hfill
     \begin{subfigure}[b]{0.14\textwidth}
         \centering
         \includegraphics[width=\textwidth]{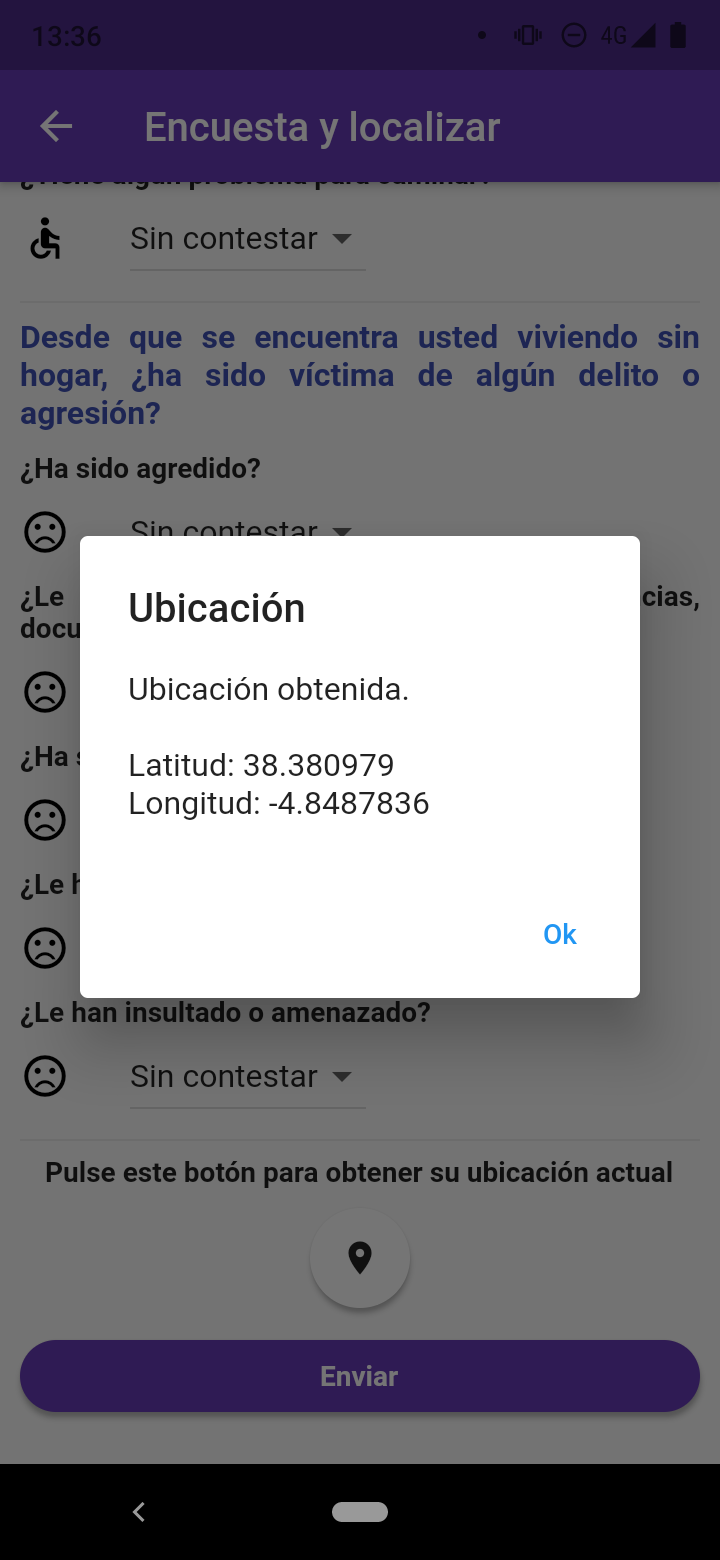}
     \end{subfigure}
    
    \begin{subfigure}[b]{0.14\textwidth}
         \centering
         \includegraphics[width=\textwidth]{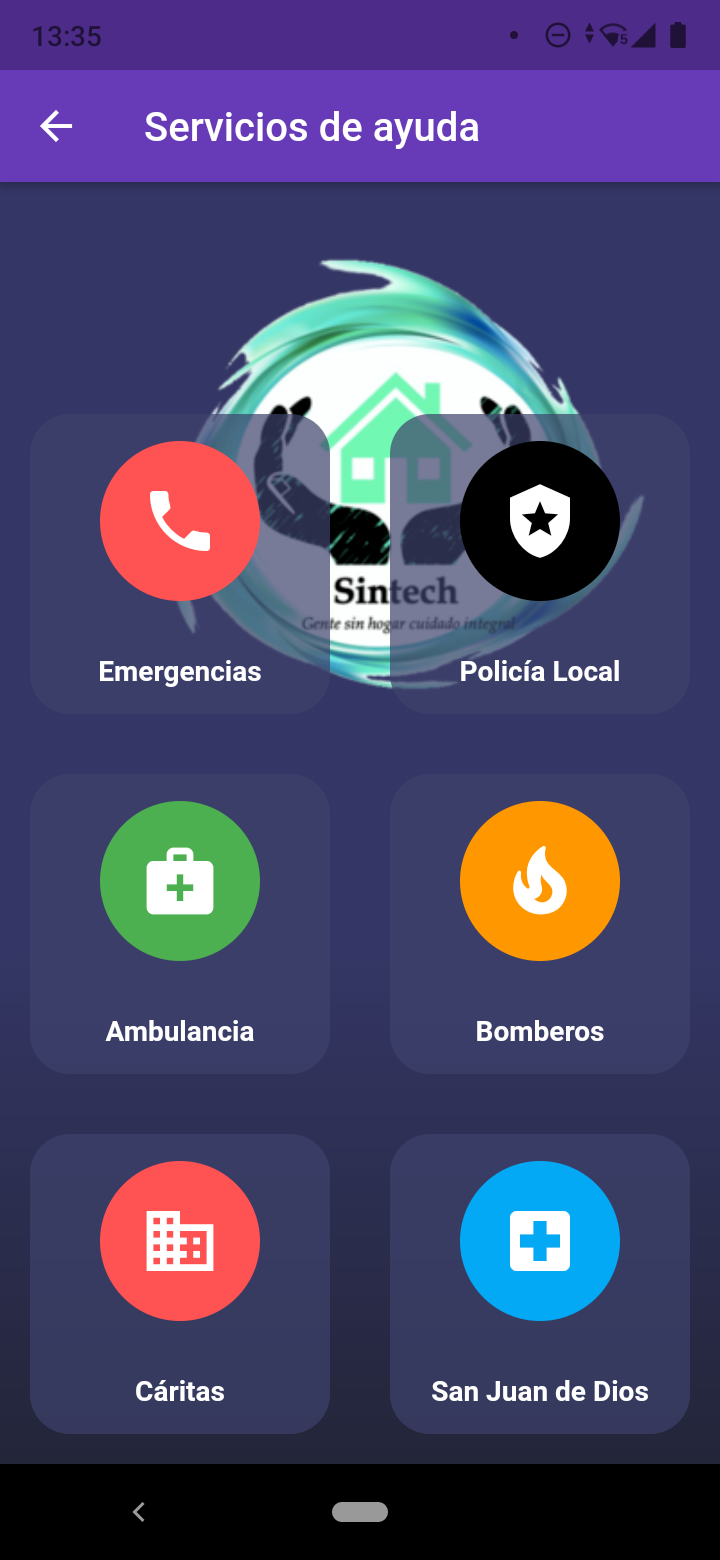}
     \end{subfigure}
     \hfill
     \begin{subfigure}[b]{0.14\textwidth}
         \centering
         \includegraphics[width=\textwidth]{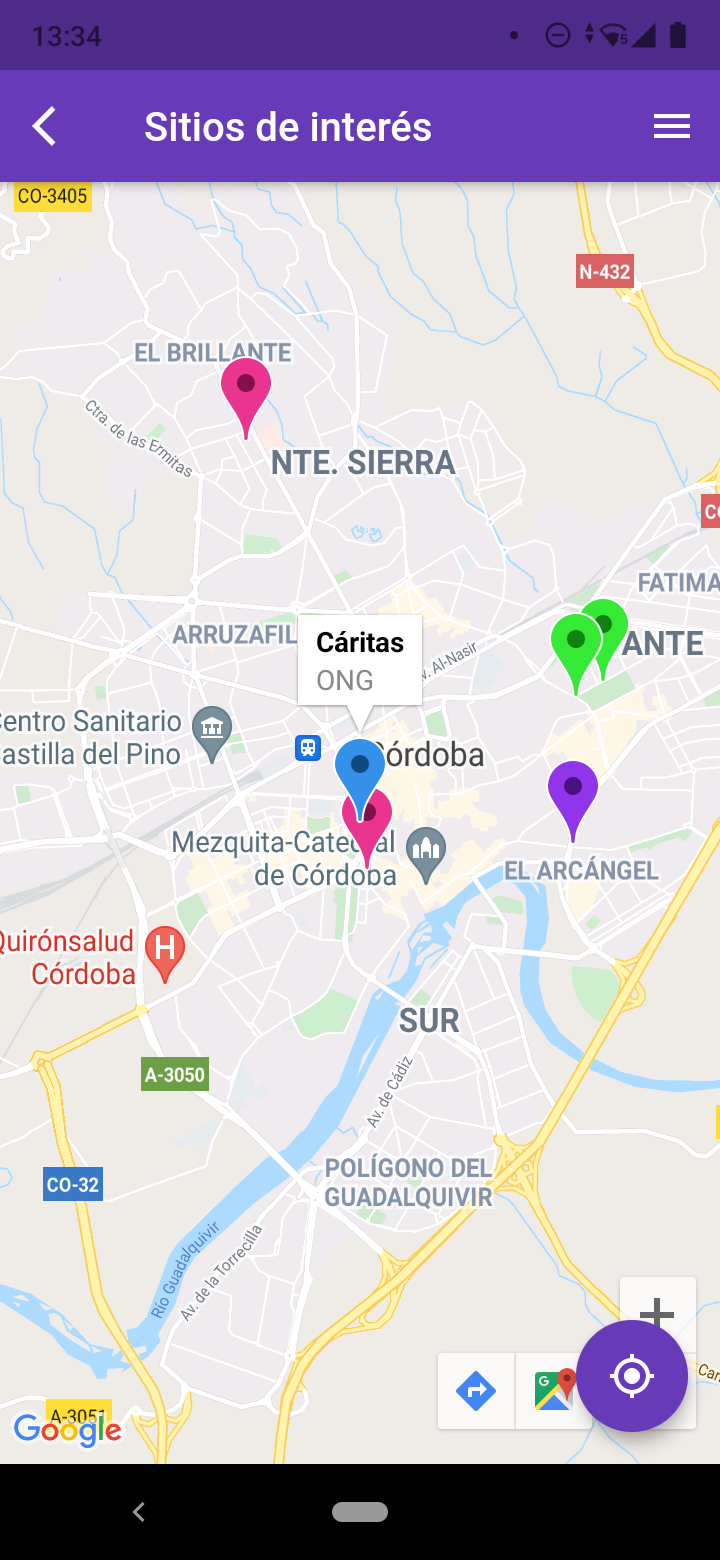}
     \end{subfigure}
     \hfill
     \begin{subfigure}[b]{0.14\textwidth}
         \centering
         \includegraphics[width=\textwidth]{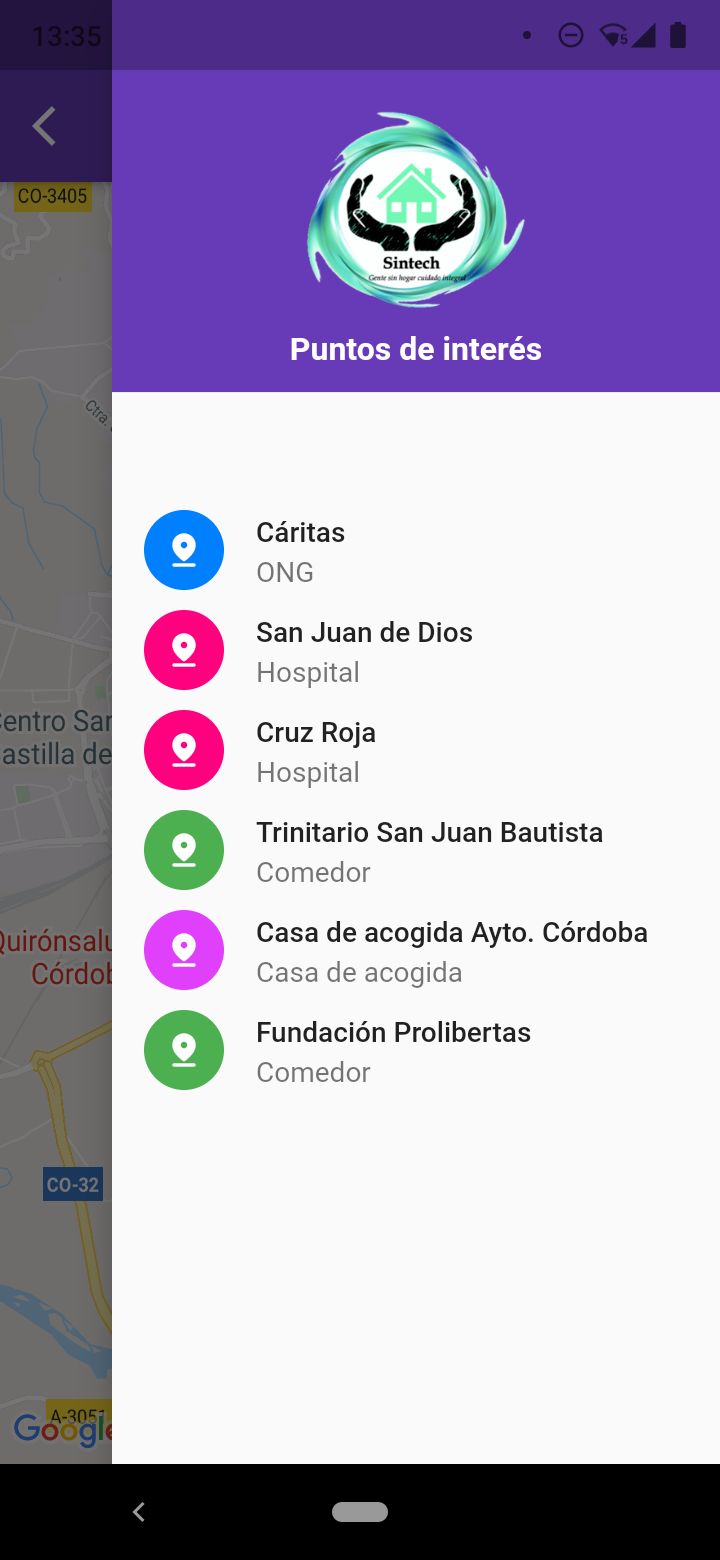}
     \end{subfigure}
        \caption{Menus and functionalities of the mobile application.}
        \label{fig:app_fotos}
\end{figure}

\subsection{Feature Selection}

During the feature selection experimentation, we compared the effectiveness of different algorithms to predict the health condition of a homeless person by using different attribute subsets. The final number of features obtained by each feature selector and the CCR resulting of applying our classifier using these features are shown in Table \ref{tab:FS_results}.

\begin{table}[h]
\caption{Classification results after applying the feature selection algorithms.}
\begin{tabular}{|c|c|c|}
\hline
\textbf{Feature Selector}                 & \textbf{Final Number of Features} & \textbf{CCR}     \\ \hline
Filter $Chi^2$               & 6                        & 87.09\% \\ \hline
Filter F-Statistic        & 6                        & 86.99\% \\ \hline
Filter Mutual Information & 6                        & 88.93\% \\ \hline
RFE                       & 3                        & 88.93\% \\ \hline
RFECV                     & 3                        & 88.93\% \\ \hline
Forward Feature Selection & 6                        & 87.77\% \\ \hline
Embedded Random Forest    & 104                      & 87.96\% \\ \hline
Random Forest as Selector & 429                      & 87.86\% \\ \hline
\end{tabular}
\label{tab:FS_results}
\end{table}

Looking at the results, we concluded that the best feature selection algorithm, according to the lower final number of features and the higher CCR score, is the RFECV. This algorithm has found a final subset of six features for which our classifier has obtained an 88.93\% CCR. However, considering the running time of the algorithm, we could consider that the best performance was achieved by the filter method using the mutual information as statistic. In this case the subset was generated 31 times faster than using the RFECV. The time comparison for applying the different feature selection algorithms is shown in the Table \ref{tab:times}.

\begin{table}[h]
\centering
\caption{Average times for feature selection algorithms.}
\begin{tabular}{|c|c|}
\hline
\textbf{Feature Selector}          & \textbf{Avg. Selection Time (s)} \\ \hline
$Chi^2$ Filter              & 0.0322                  \\ \hline
F-Statistic Filter        & 0.0311                  \\ \hline
Mutual Information Filter & 4.9325                  \\ \hline
RFE                       & 19.4441                 \\ \hline
RFECV                     & 154.1169                \\ \hline
Forward Feature Selection                       & 762.4749                \\ \hline
Embedded Random Forest    & 0.2221                  \\ \hline
\end{tabular}
\label{tab:times}
\end{table}

On the other hand, due to the possibility to select the final number of features, a more in-depth comparison has been made among the three filtering methods and the RFE algorithm. For this purpose, we performed the classification task using the different feature subsets obtained by these algorithms by varying the subset length from 1 to 100 attributes. In Figure \ref{fig:CCR_Features} the CCR evolution is shown when decreasing the number of features for the algorithms previously mentioned. The highest accuracy was achieved by the mutual information method, with an 88.93\% CCR when using 6 features. It was followed by $chi^2$ feature selector, which obtained an 88.83\% CCR when using 18 features. Then the F-statistics achieved an 88.54\% CCR when using 25 attributes. Finally, the RFE reaches an 88.44\% CCR when using 11 features. However, if we consider the mean CCR in the 100 iterations the best is the F-statistic method, followed by $chi^2$, mutual information and RFE; with CCR values of 88.10\%, 88.09\%, 88.00\% and 87.89\%, respectively. We can conclude that the highest scores were achieved by using a low number of features in all the cases, so feature selection has been useful both to improve the performance of the classifier and to reduce the complexity of the problem.

\setlength{\fboxrule}{1.2pt}
        
    

\begin{figure*}[h]
\centering
      \fbox{\includegraphics[width=0.72\textwidth]{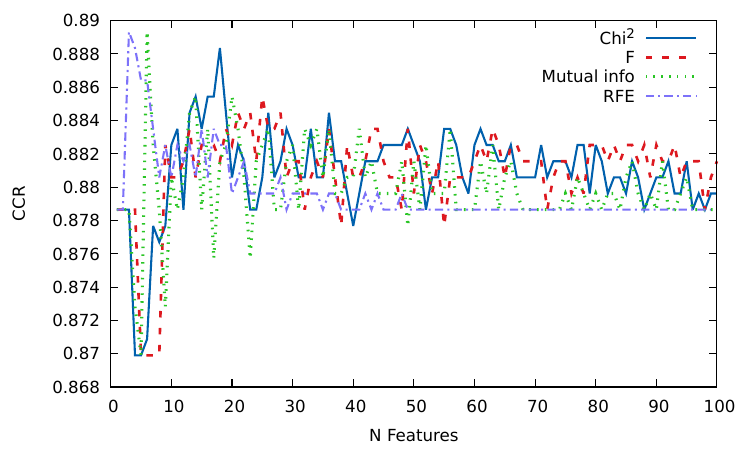}}
      \caption{Variation of the CCR when decreasing the number of features.}
      \label{fig:CCR_Features}
\end{figure*}

Finally, the most important features to predict the health state of a homeless person were determined according to the most selected attributes for the different feature selection algorithms. In Figure \ref{fig:histograma} we can see all the attributes which were selected at least by 2 of the applied algorithms when selecting their top 6 features. Looking at these results we can conclude that the most relevant variables to predict the health state of a homeless person are suffering from chronic illness, having spent any night in the hospital in the last year, and having a recognised disability.


\begin{figure}[H]
     \includegraphics[width=0.5\textwidth, center]{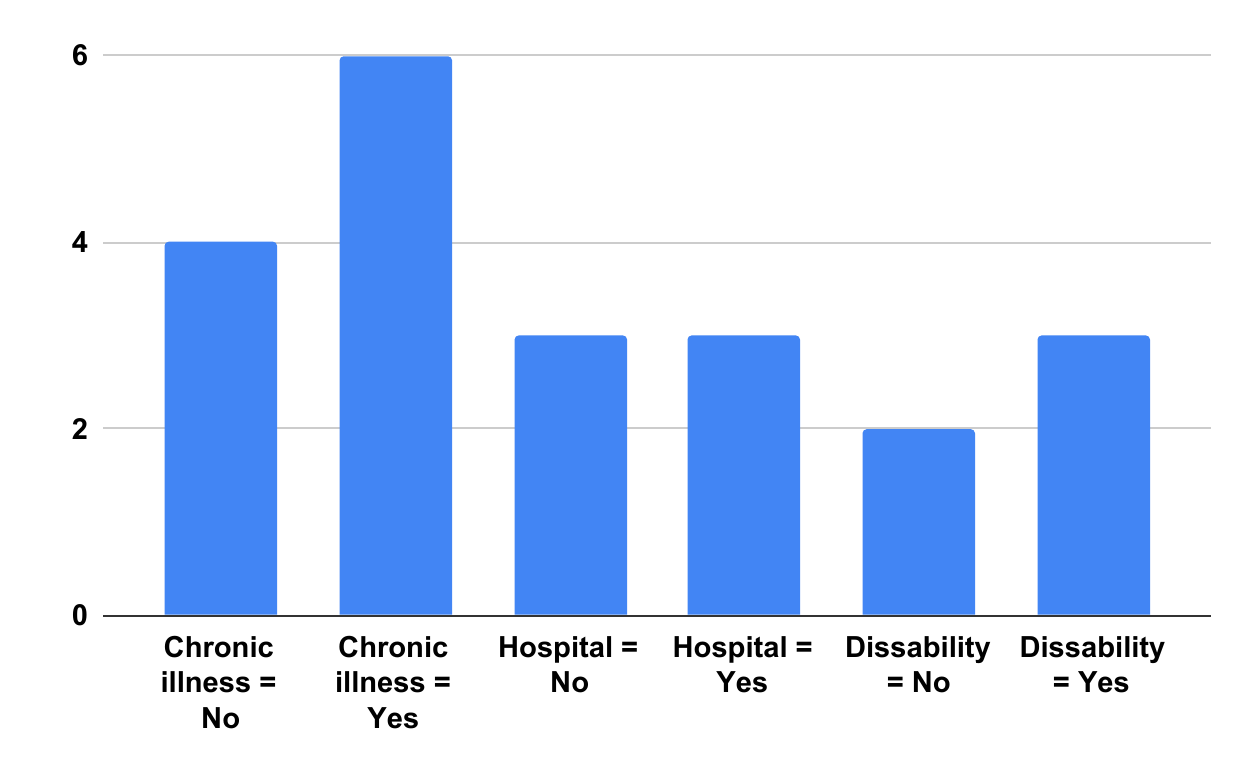}
     \caption{Most important variables to predict the health state.}
     \label{fig:histograma}
 \end{figure}

\subsection{Association Rules}

The final list of rules was obtained after applying different configurations for the algorithms. The generation of frequent itemsets by Apriori and FP-Growth was tested by variating the minimum support from 0.1 to 0.7 and the maximum itemset length from 1 to 4. After some tests, we considered as an interesting parameter configuration using a 0.1 minimum support and a 2 maximum itemset length. In this way, we obtained 38,752 itemsets as a basis for generating the association rules. Then, we started the experimentation without applying any kind of filtering and using minimum rule confidence of 0.9. That results in 10,881 association rules, which made very difficult finding interesting rules. To easy this task we perform different filterings, first using our trivial rules filter module and then according to other interest metrics: minimum lift, maximum antecedent support (M.A.S.) and maximum consequent support (M.C.S.). In Table \ref{tab:rules_exp} some of the experiments applying the different filters are shown in order to validate their efficiency. In addition, we sort the generated rules based on the lift metric to place the most interesting rules at the top of the list (according to this metric).

\begin{table}[h]
\centering
\caption{Generation and filtering of association rules.}
\begin{tabular}{|c|c|c|c|c|c|}
\hline
\textbf{Conf} & \textbf{Min Lift} & \textbf{M.A.S.} & \textbf{M.C.S.} & \textbf{Trivial Filt} & \textbf{N Rules} \\ \hline
0.9                 & 0                  & 1                          & 1                           & No                         & 10,881            \\ \hline
0.9                 & 1                  & 1                          & 1                           & No                         & 7,172             \\ \hline
0.9                 & 0                  & 0.3                        & 1                           & No                         & 3,234             \\ \hline
0.9                 & 0                  & 1                          & 0.3                         & No                         & 178               \\ \hline
0.9                 & 0                  & 1                          & 1                           & Yes                         & 9,777             \\ \hline
0.9                 & 1                  & 0.5                        & 0.5                         & No                         & 516               \\ \hline
0.9                 & 1                  & 0.5                        & 0.5                         & Yes                         & 85                \\ \hline
0,7                 & 0                  & 1                          & 1                           & No                         & 26,526            \\ \hline
0,7                 & 1                  & 0.5                        & 0.5                         & No                         & 766               \\ \hline
0,7                 & 1                  & 0.5                        & 0.5                         & Yes                         & 293               \\ \hline
\end{tabular}
\label{tab:rules_exp}
\end{table}

The lists of association rules generated after the different experiments were analysed to find rules which contain interesting information about the homeless people. Some of these interesting rules are shown in Table \ref{tab:rules}.

\begin{table}[h]
\caption{Interesting association rules.}

\begin{tabular}{|c|c|}
\hline
\textbf{Association Rules}                                                                  & \textbf{Confidence} \\ \hline
Recognised Disability == Yes $\xrightarrow{}$ Nationality == Spanish     & 90.60\%       \\ \hline
Age == {[}18-28{]} $\xrightarrow{}$ Nationality ==  Foreign              & 84.77\%       \\ \hline
Age == {[}28-37{]} $\xrightarrow{}$ Nationality ==  Foreign              & 72.21\%       \\ \hline
Arrested == Yes, many times $\xrightarrow{}$ Drugs consume == Yes & 74.70\%       \\ \hline
Arrested == Yes, many times $\xrightarrow{}$ Nationality == Spanish & 71.95\%       \\ \hline
\end{tabular}
\label{tab:rules}
\end{table}

By looking at these rules, we can obtain some information about homelessness that may be of interest to NGOs:

\begin{itemize}
    \item Foreigners with disabilities, if any, usually do not have a recognised disability.
    \item Around 80\% of homeless people under the age of 37 are not Spanish, with a higher tendency for those under 28.
    \item Drug consumption is present in almost 75\% of homeless people who have been arrested on multiple occasions. Moreover, these people tend to be Spanish. It should be noted that this does not imply that drug consumption results in the arrest of the person, but that it is common among those who have been arrested.
\end{itemize}

Finally, we obtained some class association rules by introducing the developed grammar in the PonyGE2 \cite{ponyge2} algorithm and taking the health state as objective variable to predict. Some of these generated rules are shown in Table \ref{tab:class_rules}. Most of the resulting rules are trivial and have little relevance to the study. However, we have proved the effectiveness of our grammar to generate class association rules by using a genetic algorithm and a custom interest metric. Therefore, by refining the grammar and metric of interest we could obtain a good class association rule generator based on a genetic algorithm, in which we can define the structure of the rules we want to be generated.

\begin{table}[!h]
\caption{PonyGE2 class association rules}

\begin{tabular}{|c|c|}
\hline
\textbf{Class Association Rules}                                                                  & \textbf{Confidence} \\ \hline
Chronic Disease == No $\xrightarrow{}$ Health State == Good     & 90.60\%       \\ \hline
\begin{tabular}{@{}c@{}}Recognised Disability != Yes \textit{AND} \\ Chronic Disease == No $\xrightarrow{}$ Health State == Good\end{tabular}              & 84.77\%       \\ \hline
\end{tabular}
\label{tab:class_rules}
\end{table}

\section{Conclusions and Future Work}
In this paper, we have proposed three different tools to support NGOs in their work to help homeless people. The mobile application is already being used to collect new information on homeless people. Feature selection has not only shown us the importance of reducing the dimensionality of the problem in order to construct classifiers, but has also obtained the most important features for predicting the health state of a homeless person. Finally, the association rules module have shown some interesting relationships among the attributes of homeless people, as well as another way of predicting their health state based on specific characteristics using an evolutionary algorithm. In the future, we will improve some of the proposed methods, especially the evolutionary generation of class association rules, and we will apply them to our application collected data for an optimum homeless data analysis and support to NGOs work.


\section*{Acknowledgement}
The authors would like to thank all members of the project ``SINTECH: Tecnología para el apoyo a las personas sin hogar'' and the Spanish non profit organisation ``C\'aritas Diocesana'' for their contribution and support to this work. This research has been funded by the Ministry of Education and Professional Training (Collaboration Scholarship, Code: 998142). Additionally, this research was framed on the UCO SOCIAL INNOVA VI from the University of Cordoba (Reference: 2020/00757).

\bibliographystyle{ieeetr}
\bibliography{bibliography}\clearpage

\end{document}